\pdfoutput=1

\documentclass[11pt]{article}

\usepackage[]{ACL2023}

\usepackage{times}
\usepackage{latexsym}

\usepackage[T1]{fontenc}

\usepackage[utf8]{inputenc}

\usepackage{adjustbox}

\usepackage{multirow}
\usepackage{soul} 
\usepackage{multicol}
\usepackage{diagbox}
\usepackage{booktabs}
\usepackage{graphicx}
\usepackage{subcaption}
\usepackage{graphics}
\usepackage{xspace}
\usepackage{amsmath}
\usepackage{amssymb}  

\newcommand{\dataName}[0]{$\hat{P}aRTE$\xspace}
\newcommand{\verbose}[1]{}

\usepackage{microtype}

\usepackage{inconsolata}

\title{Evaluating Paraphrastic Robustness in Textual Entailment Models}

\author{Dhruv Verma \quad\quad
  Yash Kumar Lal
  \\ Stony Brook University \\ \texttt{\{dhverma,ylal\}@cs.stonybrook.edu} \And 
  Shreyashee Sinha \\
  Bloomberg \\\texttt{ssinha176@bloomberg.net}
  \AND
  Benjamin Van Durme \\
  Johns Hopkins University
  \\\texttt{vandurme@jhu.edu} \And 
  Adam Poliak \\
  Bryn Mawr College \\\texttt{apoliak@brynmawr.edu} }

\begin{document}
\maketitle

\begin{abstract}
We present \dataName, a collection
of 1,126 pairs of Recognizing Textual Entailment
(RTE) examples to evaluate whether 
models are robust to paraphrasing. 
We posit that if RTE models understand language, their predictions should be consistent across inputs that share the same meaning.
We use the evaluation set to determine if RTE models' predictions change when examples are paraphrased.
In our experiments, 
contemporary models 
change their predictions on 8-16\% of paraphrased examples, indicating that there is still room for improvement. 

\end{abstract}

\section{Introduction}

Recognizing Textual Entailment (RTE), the task of predicting whether one sentence (\textit{hypothesis}) would likely  be implied by another (\textit{premise}),
is 
central to natural language understanding \cite[NLU;][]{dagan2005pascal},
as this task captures
``all manners of linguistic phenomena and
broad variability of semantic expression''~\cite{maccartney2009natural}.
If an RTE model has a sufficiently high \textit{capacity for reliable, robust inference necessary for full NLU}~\cite{maccartney2009natural}, then the model's predictions should be consistent 
across paraphrased examples.

We introduce \dataName, a test set to evaluate how \textit{reliable} and \textit{robust} models are to paraphrases (\autoref{tab:nli_examples} includes an example).
The test set 
consists of examples from the Pascal RTE1-3 challenges~\cite{dagan2006pascal,BarHaim2006TheSP,giampiccolo2007third} rewritten with a lexical rewriter and manually verified to preserve the meaning and label of the original RTE sentence-pair. 
We use this evaluation set to determine whether models
 change their predictions when examples are paraphrased.

While this may not be a sufficient test to determine whether RTE models \textit{fully understand} language, as there are many semantic phenomena that RTE models should capture~\cite{cooper1996using,naik-EtAl:2018:C18-1}, 
 it is \textit{necessary} that any NLU system be robust to paraphrases.

Our experiments indicate that contemporary models are robust to paraphrases 
as their predictions do not change on the overwhelmingly large majority of examples that are paraphrased.
However, our analyses temper this claim as models are more likely to change their predictions when both the premise and hypothesis are phrased compared to when just one of the sentences is rewritten.
We release \dataName\footnote{\url{https://dataverse.harvard.edu/dataset.xhtml?persistentId=doi:10.7910/DVN/HLMI23}} to encourage others to evaluate how well their models perform when RTE examples are paraphrased.

\begin{table}[t!]
 \centering
 \small
 \renewcommand{\arraystretch}{1.25}
 \begin{tabular}{p{.3cm}p{6.5cm}}
     \midrule
 \textbf{P} & The cost of security when world leaders gather near Auchterarder for next year 's G8 summit, is expected to top \$150 million. \\
  \textbf{P'} & The cost of security when world leaders meet for the G8 summit near Auchterarder next year will top \$150 million.\\
  \hline 
  \textbf{H} & More than \$150 million will be probably spent for security at next year's G8 summit. \\
 \textbf{H'} & At the G8 summit next year more than \$150 million will likely be spent on security at the event. \\ 
\bottomrule
 \end{tabular}
 \caption{An original and paraphrased RTE example. 
 The top represents an original premise (P) and its paraphrase (P'). The bottom depicts an original hypothesis (H) and its paraphrase (H'). 
 A model robust to paraphrases should have consistent predictions across the following pairs: P-H, P'-H, P-H', and P'-H'.}
 \label{tab:nli_examples}
 \end{table}

\section{Related Work}

With the vast adoption of human language technology (HLT), systems must understand when different expressions convey the same meaning (paraphrase) and support the same inferences (entailment).
Paraphrasing and entailment are closely connected as the former is a special
case of the latter where two sentences entail each other~\cite{nevvevrilova2014paraphrase,fonseca-aluisio-2015-semi,vita2015computing,ravichander-et-al-2022-condaqa}.
Paraphrasing has been used to improve RTE predictions~\cite{bosma2006paraphrase,sun-etal-2021-aesop} and RTE has been used for paraphrase identification~\cite{8342717} and generation~\cite{AMA_paper}.
Furthermore, both phenomena are key to NLU~\cite{androutsopoulos2010survey} and work such as \citet{zhao2018generating,hu-etal-2019-large} have explored rewriting RTE examples to create more robust models.

We follow a long tradition of evaluating linguistic phenomena captured in RTE models~\cite{cooper1996using}.
Recent tests focus on evaluating how well contemporary RTE models capture phenomena such as
monotonicity~\cite{yanaka2019can,yanaka2019help}, verb veridicality~\cite{ross-pavlick-2019-well,yanaka-etal-2021-exploring}, presuppositions~\cite{parrish-etal-2021-nope} implicatures~\cite{jeretic-etal-2020-natural}, basic logic~\cite{Richardson2020ProbingNL,shi-etal-2021-neural}, figurative language~\cite{chakrabarty-etal-2021-figurative}, and others~\cite{naik-EtAl:2018:C18-1,poliak-etal-2018-collecting-diverse,vashishtha-etal-2020-temporal}.
Unlike many of those works that evaluate models' accuracy on examples that target specific phenomena, we use a contrastive approach~\cite{prabhakaran-etal-2019-perturbation,gardner-etal-2020-evaluating} to determine whether RTE models' predictions change when
examples are paraphrased.

\section{\dataName}

 To explore whether these RTE models are robust to paraphrases, we
create \dataName, a modified version of 
the Pascal RTE1-3 challenges~\cite{dagan2005pascal,BarHaim2006TheSP,giampiccolo2007third}.
\dataName contains 1,126 examples of an original unmodified RTE sentence-pair grouped with a sentence-pair with a modified premise, hypothesis, or both.
We use the examples in RTE1-3 to create our test set, as opposed to other RTE datasets
due to its long-standing history.

\subsection{Paraphrase Generation \& Verification}

For each RTE premise-hypothesis pair (P-H), we created three paraphrased premises (P') and 
hypotheses (H') using a T5-based paraphraser\footnote{We manually verified the quality of this paraphraser. See \autoref{app:dataset-crowd} for more details.} fine-tuned on the Google PAWS dataset~\cite{zhang2019paws}.
To ensure lexically diverse paraphrases, we filter out any paraphrases that have high lexical overlap with the original sentences using Jaccard index threshold of 0.75.
Out of 14,400 generated sentences, 2,449 remained - 956 paraphrased premises (P') and 1,493 paraphrased hypotheses (H').
Next, we retained 
550 paraphrased premises and 800 paraphrased hypotheses paraphrases
that crowdsource workers identified as grammatical and similar in meaning to the original sentences.\footnote{See \autoref{app:dataset-crowd} for a detailed description of this filtering process, including annotation guidelines.}
We include a grammatical check since an existing RTE evaluation set focused on paraphrases~\cite{white-EtAl:2017:I17-1} contains hypothesis-only biases related to grammaticality~\cite{hypoths-nli}.

 If at least one P' or one H' passes this filtering process, we retain the original RTE example
and pair it with a corresponding paraphrased example (i.e. P'-H', P'-H, or P-H').
In the case where more than one P' or H' passes the filtering,
we retained the P' or H' that  crowdsource workers deemed most similar to the original sentence.
Out of the original 2,400 RTE test pairs, we retain 914 pairs with a high-quality P' or H', 
resulting in 1,178 original and paraphrased RTE pairs.\footnote{415 pairs where the premise is paraphrased, 631 pairs where the hypothesis is paraphrased, and 132 pairs where both are paraphrased.}

\begin{table*}[t!]
\centering
\begin{tabular}{|p{2.5cm}|p{2.5cm}|p{2.5cm}|p{2.5cm}||p{2.5cm}|}
\hline
 \multirow{2}{*}{\diagbox[width=7.5em]{Model}{Testset}} & \multirow{2}{*}{MNLI} & \multirow{2}{*}{RTE} & \multirow{2}{*}{\dataName}  & \multirow{2}{*}{\% $\Delta$ \dataName} \\
 & & & & \\
 \hline
BoW      &  67.97   &   53.99  &  54.70 &  15.27   \\ \hline 
BiLSTM & 66.68   &   51.59  &  51.24  &      16.69 \\ \hline
BERT   & 90.04   &   72.11  &  72.55 &  9.50    \\ \hline
RoBERTa   & 92.68   & 
83.83  & 82.59  &  7.99    \\ \hline 
GPT-3   & -   &   80.90  & 79.12 &  10.12  \\ \hline
\end{tabular}
\caption{Each row represents a model. The columns MNLI, RTE, \dataName report the model’s accuracy on those test sets.  The last column (\% $\Delta$ \dataName) reports the percentage of examples where the model changed its prediction.
}
\label{table:resultsall}
\end{table*}

\subsection{Overcoming Semantic Variability}
\citet{maccartney2009natural} argues that in addition to being \textit{reliable} and \textit{robust}, RTE models must deal with the \textit{broad variability of semantic expression}. In other words,
though two sentences may be semantically congruent, it is possible that small variations in a paraphrased sentence contain enough semantic variability to change what would likely, or not likely be inferred from the sentence. 
Despite all P' and H' being deemed to be semantically congruent with their corresponding original sentences, the semantic variability of paraphrases might change whether H or H' can be inferred from P' or P.

Therefore, propagating an RTE label from an original sentence pair to a modified sentence pair might be inappropriate.
We  manually determined that this issue occurs in just  52 (4\%) 
examples,
and retained 
1,126 examples. This ensures an evaluation set of high-quality examples that can be used to determine whether models are sensitive to paraphrases and 
change their prediction on paraphrased examples.
Our dataset contains 402 examples with just a paraphrased premise P', 602 with just a paraphrased hypothesis H', and 122 with both a paraphrased premise and hypothesis.

\section{Experimental Setup}

We explore models built upon three different classes of sentence encoders: bag of words (BoW), LSTMs, and Transformers.
Our BoW model represents premises and hypotheses as an average of their tokens' 300 dimensional 
GloVe embeddings~\cite{pennington2014glove}. 
The concatenation of these representations is fed to an MLP with two hidden layers.
For the BiLSTM model,
we represent tokens with GloVe embeddings, extract sentence representations using max-pooling, and pass concatenated sentence representations  to an MLP with two hidden layers.

Our transformer-based models are pre-trained 
BERT~\cite{devlin-etal-2019-bert} and Roberta~\cite{liu2020roberta} encoders with an MLP attached to the final layer.
Additionally, we use GPT-3 in a zero-shot setting where we ask it  
 to label the relationship between a premise and hypothesis.\footnote{See \autoref{app:implementation} for more details, including hyper-parameters, model sizes, and GPT-3 prompt design and configurations. Our code is available at \url{https://github.com/stonybrooknlp/parte}}

The RTE training sets do not contain enough examples to train deep learning models with a large number of parameters. We follow the common practice of training models on MNLI and using our test set to evaluate how well they capture a specific phenomenon related to NLU.
During testing, 
we map the MNLI `contradiction' and `neutral' labels to the `not-entailed' label in RTE, following common practice~\cite[][\textit{inter ailia}]{
wang2018glue,yin-etal-2019-benchmarking,ma-etal-2021-issues,utama-etal-2022-falsesum}.

\begin{figure}[!t]
    \centering
     \adjustbox{width=\columnwidth}{
    \includegraphics[]{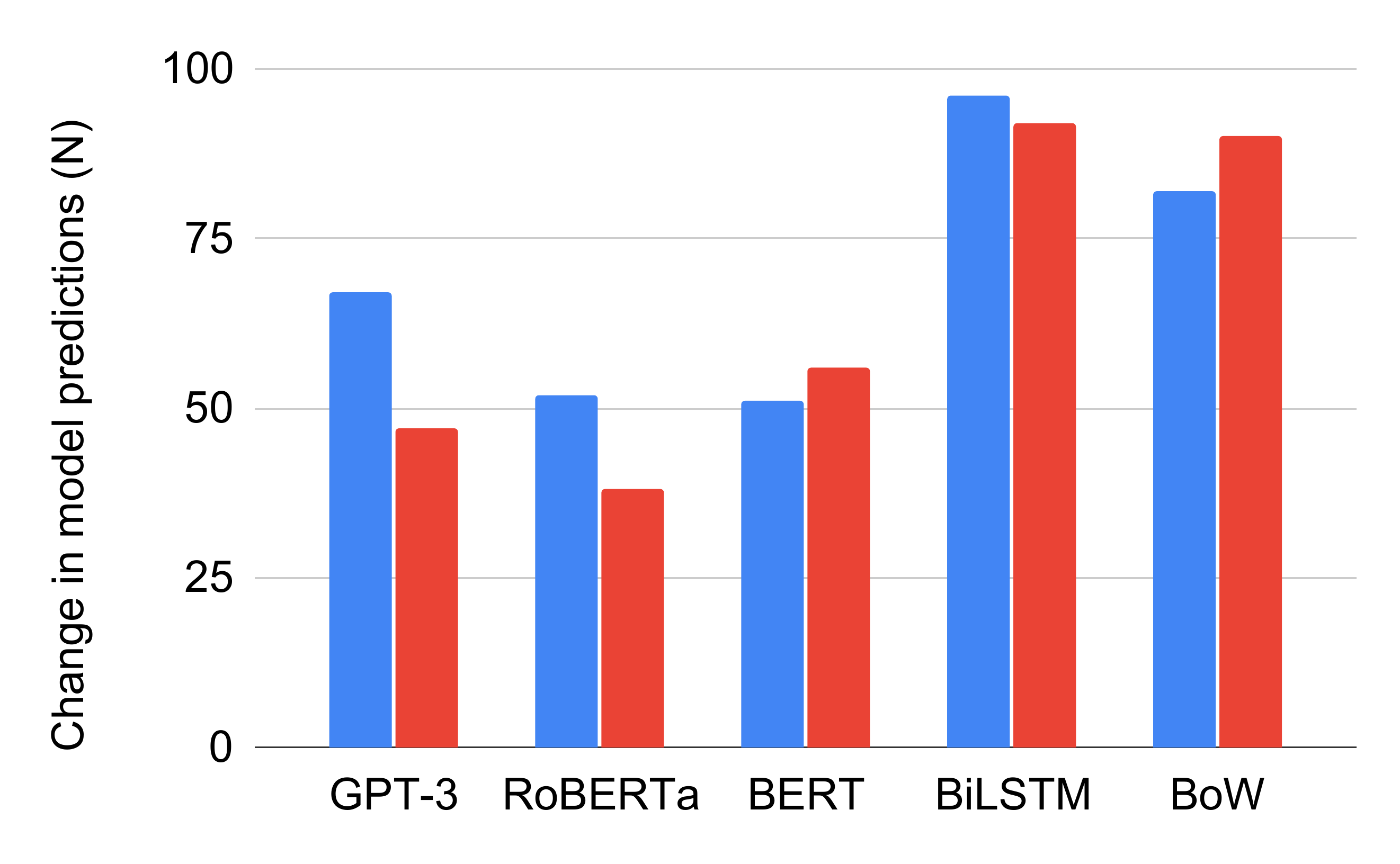}}
    \caption{
    Number of times a model changes its predictions from correct to incorrect (left blue bar) or incorrect to correct (right red bar).}
    \label{fig:incorrect_correct_model_changes}
\end{figure}

\section{Results}

\autoref{table:resultsall} report the results.
The RTE and \dataName columns respectively report the models' accuracy on the 1,126 unmodified and paraphrased sentence pairs.\footnote{Although there are just 914 unmodified sentence pairs, for the sake of a head-to-head comparison, we retain all instances of the unmodified sentence pairs when computing accuracy.}
Comparing the difference in accuracy between unmodified and paraphrased examples can be misleading. 
If the number of times a model changes 
a correct prediction is close to the number of times it changes an incorrect prediction,
then the accuracy will hardly change. 
\autoref{fig:incorrect_correct_model_changes} demonstrates why the accuracies do not change by much when models' predictions change on paraphrased examples.
Furthermore, if a model is robust to paraphrases, then it should not change its predictions when an example is paraphrased, even if the prediction on the original unmodified example was incorrect.
Hence, our test statistic is the percentage of examples where a model's predictions change  (\% $\Delta$ \dataName column in \autoref{table:resultsall}) rather than a change in accuracy.

Compared to the Transformer based models, the  BoW and BiLSTM models seem to be more sensitive, and less robust to paraphrasing, as they change their predictions on 15.27\% and 16.69\% respectively of the 1,126 examples. However, this might be associated with how word xembedding models only just outperform random guesses in and perform much worse on RTE compared to the Transformer models.

Focusing on the Transformer models,
we noticed that RoBERTa performs the best on the datasets and is the most robust to paraphrasing - changing its predictions on just under 8\% of paraphrased examples. 
Interestingly, when the models are trained specifically to perform this task, the models change their predictions on fewer paraphrased examples as these models' accuracy increases. However, improving performance alone might not automatically improve models' robustness to paraphrases. GPT-3's accuracy noticeably outperforms BERT's accuracy, but GPT-3 changes its predictions on more paraphrased examples compared to BERT.  

\begin{figure}[!tb]
    \centering
    \adjustbox{width=\columnwidth}{
    \includegraphics[]{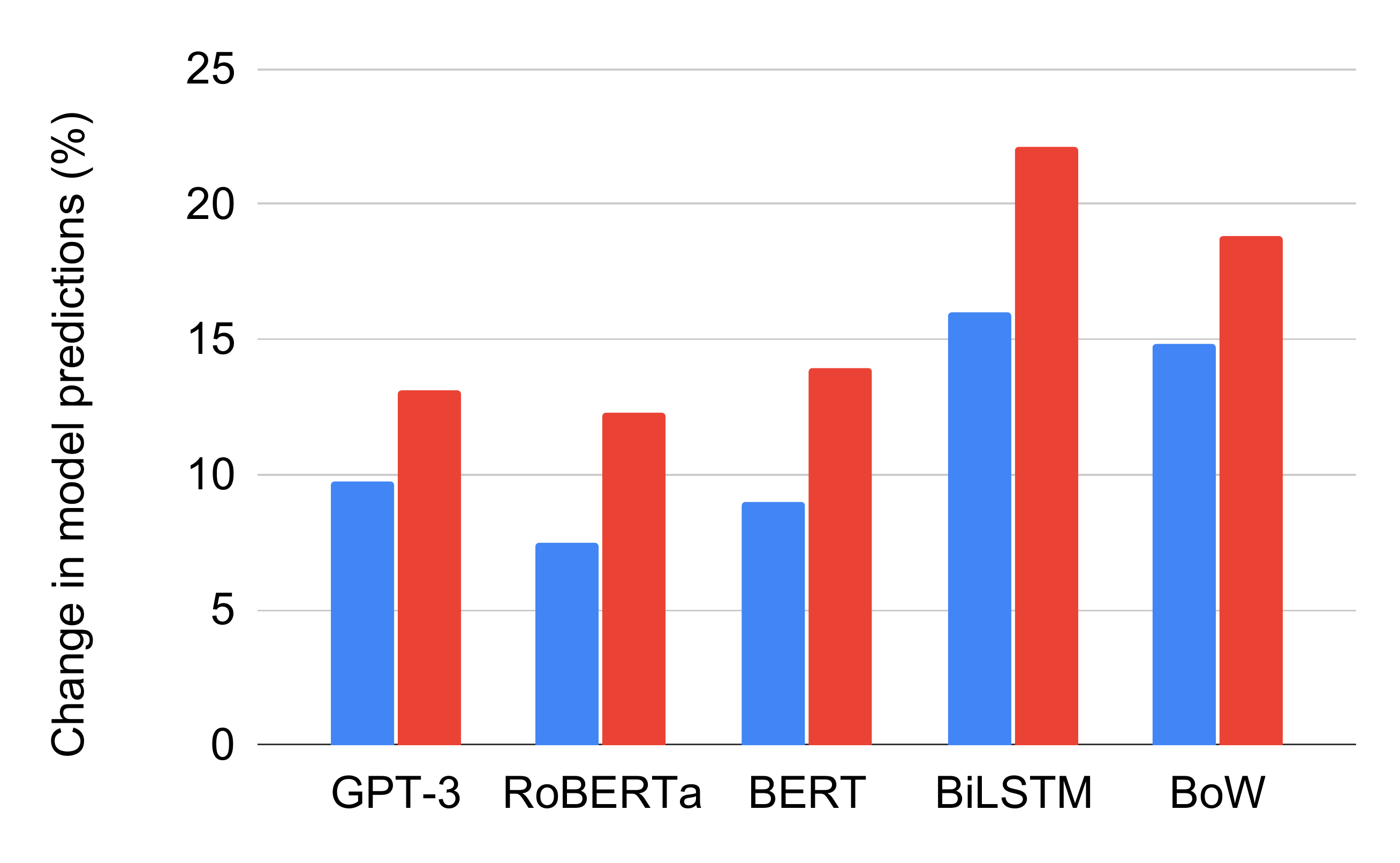}}
    \caption{Percentage of examples with one paraphrased sentence (left blue bar) or two paraphrased sentences (right red bar) where models' predictions change.}
    \label{fig:para_consistency}
\end{figure}

\paragraph{P'-H' compared to P-H' or P'-H}
\autoref{fig:para_consistency} shows noticeable increases in the percentage of changed predictions when both premise and hypothesis are paraphrased compared to when just one of the sentences is paraphrased. Specifically, for BoW and BiLSTM we see an increase of 4.01 and 6.01 percentage points respectively, and for BERT, Roberta, GPT-3 increases of 4.97, 4.83, and 3.55. As the transformer-based models changed their predictions on 12-14\% of examples where both sentences are paraphrased compared to 9-11\% in general, this analysis 
further suggests that these models are not as robust to paraprhases as desired.

\begin{figure}[!tb]
    \centering
        \adjustbox{width=\columnwidth}{
    \includegraphics[]{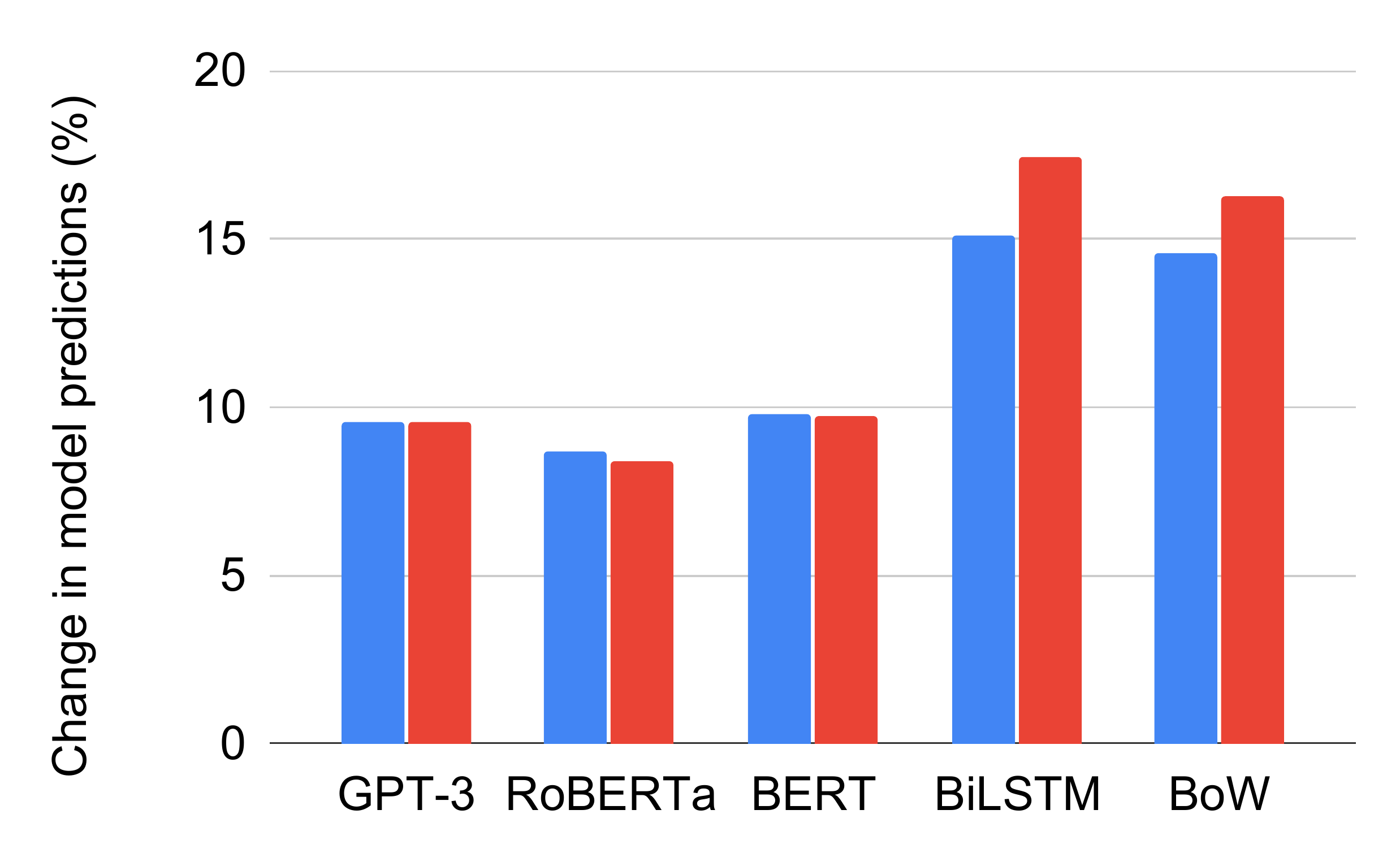}}
    \caption{Percentage of examples where the models' predictions changed when the gold label is entailed (blue left bar) or not-entailed (right red bar).}
    \label{fig:consistency_by_label}
\end{figure}

\paragraph{Entailed vs Not-entailed examples}
RTE analyses often differentiate how models perform on entailed vs not entailed examples~\cite{liu-etal-2022-wanli}.
In \autoref{fig:consistency_by_label}, we do not see meaningful differences in how models' predictions change on paraphrased examples based on the gold label. 
This might suggest that our dataset does not contain   statistical irregularities based on the 
RTE labels.

\begin{figure}[!tb]
    \centering
        \adjustbox{width=\columnwidth}{
    \includegraphics[]{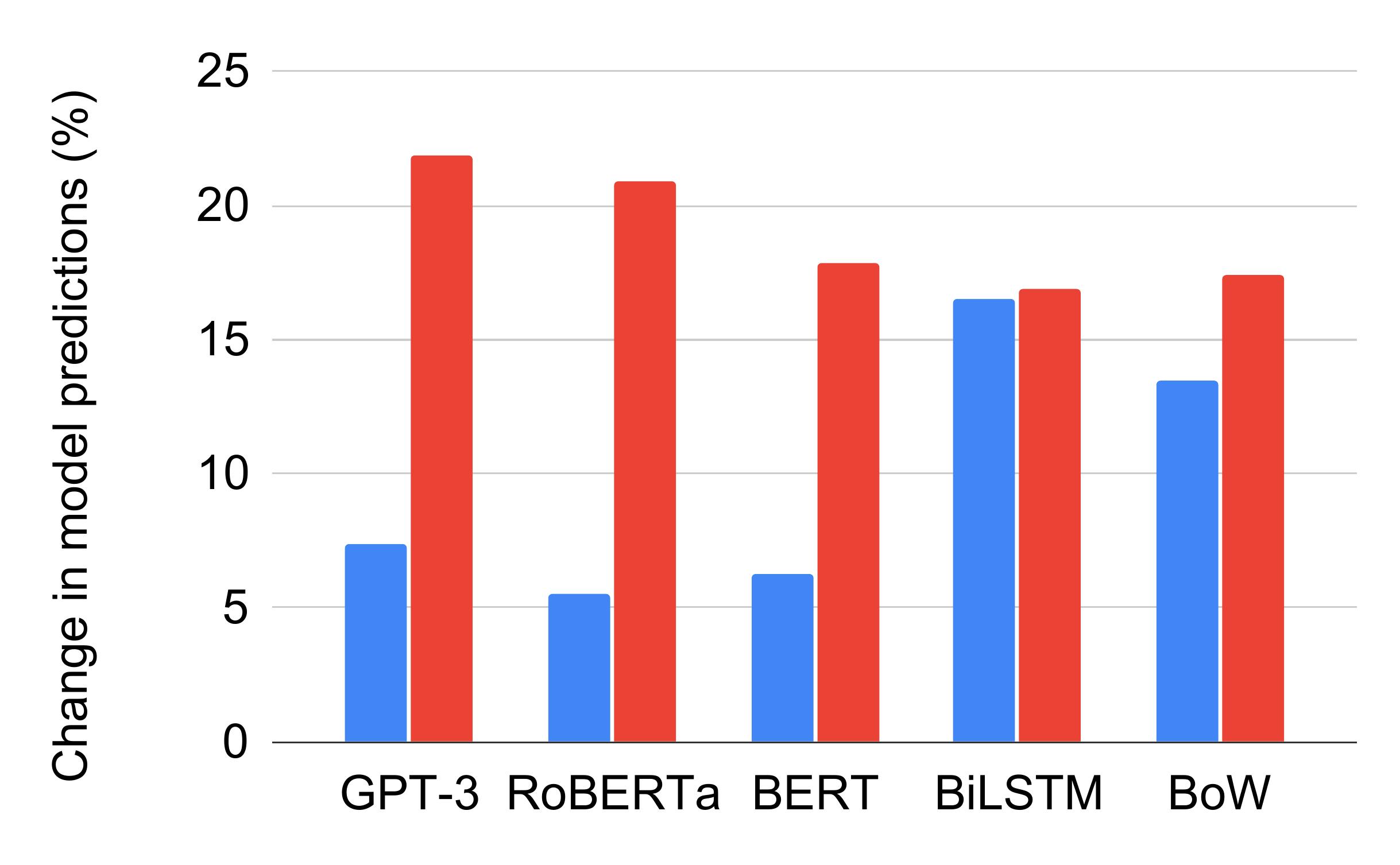}}
    \caption{Percentage of examples where models' predictions change when the original prediction was correct (left blue bar) or incorrect (right red bar).}
    \label{fig:consistency_by_preds}
\end{figure}

\paragraph{Correct vs Not-Correct Predictions}

\autoref{fig:consistency_by_preds} shows that the Transformer models' predictions is more likely to change when it's prediction on an original example was incorrect (right red bars) compared to when the prediction for an original example was correct (left blue bars). 
For example,
when RoBERTa's prediction for an original RTE example was correct,  
the model changed its prediction on 
just 5.5\% of the corresponding paraphrased examples. When RoBERTa's predictions for an original RTE example were incorrect, RoBERTa's predictions changed for 20.88\% corresponding paraphrased examples. Analyzing differences in models' confidences assigned to predictions might provide more insight~\cite{marce-poliak-2022-gender}. We leave this for future work. 

\begin{figure}[!t]
    \centering
     \adjustbox{width=\columnwidth}{
    \includegraphics[]{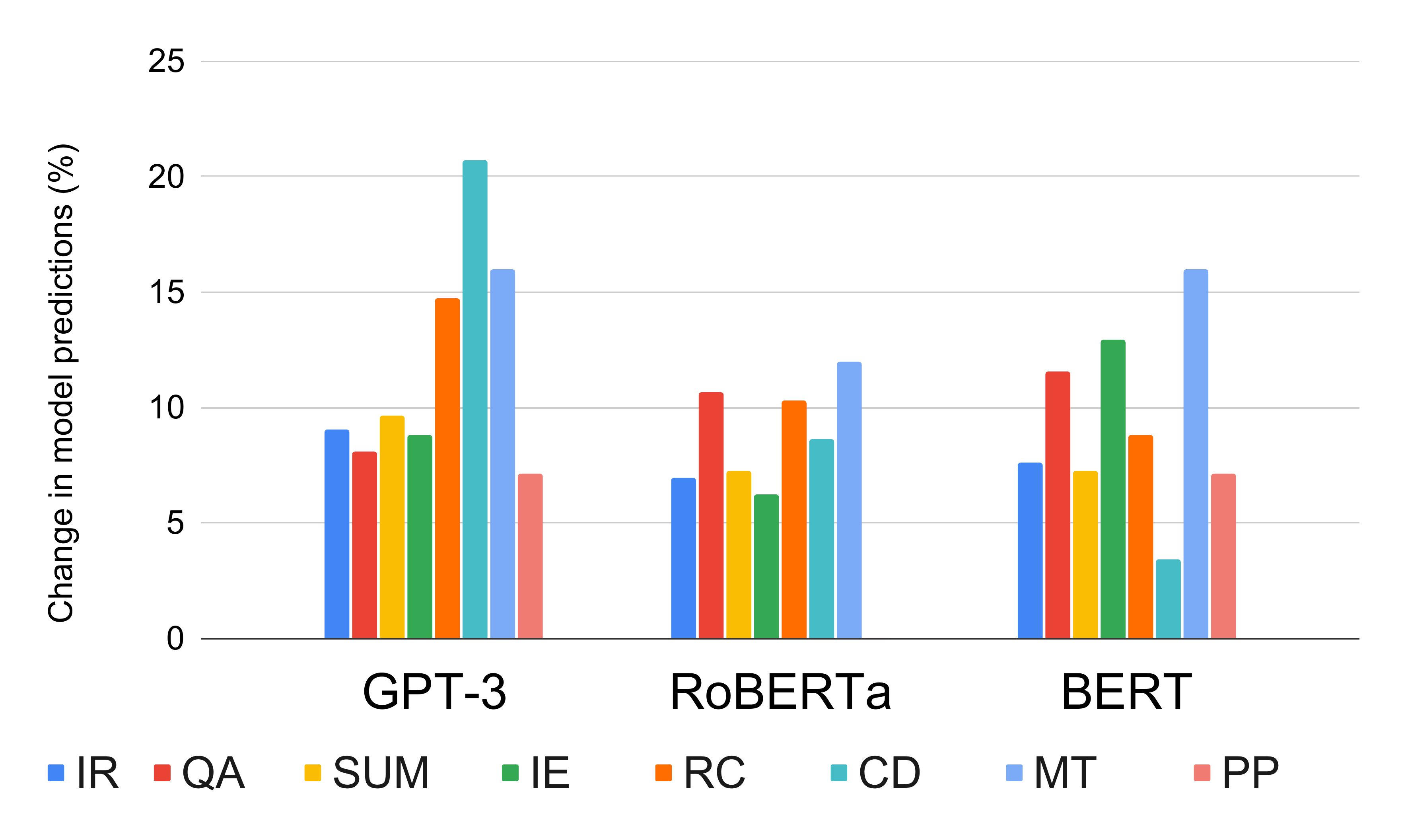}} 
    \caption{
    Percentage of examples where models predictions change their predictions depending on the examples' sources. We omit BoW and BiLSTM for space.} 
    \label{fig:inconsistency_by_task}
\end{figure}

\paragraph{Source Task}
RTE1-3 examples originated from multiple domains and downstream tasks, e.g. question-answering~\cite{Moldovan2006ATP}, information extraction~\cite{grishman-sundheim-1996-message}, and summarization~\cite{evans-etal-2004-columbia,radev-etal-2001-newsinessence}. This enables researchers to evaluate how RTE models perform on examples that contain different
aspects of \textit{open domain inference} necessary for the task~\cite{maccartney2009natural}.
\autoref{fig:inconsistency_by_task} reports the changes in models' predictions across the different sources of examples.
 We do not see consistent trends across
 the original data sources. 

\section{Conclusion}

We introduced \dataName, a high-quality evaluation set of RTE examples paired with paraphrased RTE examples. We use our evaluation set to determine whether RTE models are robust to paraphrased examples. 
Our experiments indicate that while these models predictions are usually consistent when RTE examples are paraphrased, there is still room for improvement as models remain sensitive to changes in input~\cite{jia-liang-2017-adversarial,belinkov2018synthetic,iyyer-etal-2018-adversarial}. We hope that researchers will use \dataName to evaluate how well their NLU systems perform on paraphrased data.

\section*{Limitations}

Our results nor evaluation set cannot be used to indicate whether RTE models trained for other languages are robust to paraphrases. However, researchers can apply the methods we used to develop \dataName to build evaluation sets in other languages to test whether non-English NLU systems are robust to paraphrases. 

\section*{Ethics Statement}
In conducting our research on RTE model robustness to paraphrasing, we take great care to ensure the ethical and responsible use of any data and models involved. We adhere to the principles of fairness, transparency, and non-discrimination in our experimentation and analysis. Furthermore, we take measures to protect the privacy and confidentiality of any individual crowdsource workers.
We also strive to make our evaluation set and methods openly available to the research community to promote further study and advancement in the field of Natural Language Processing. 

\bibliography{anthology,custom}
\bibliographystyle{acl_natbib}

\clearpage

\appendix

\section{Experimental Implementation Details}
\label{app:implementation}

This section describes the model implementations for our experiments. For our work we trained/fine-tuned three different models - Bag of Words (BoW), BiLSTM, BERT-large with a classification head and RoBERTa-large with a classification head. 
Each model was trained on the MultiNLI training dataset~\cite{williams-etal-2018-broad} and validated on the paraphrased RTE dev set we created. 
Each model was implemented using PyTorch.
All transformer based models were downloaded from HuggingFace.

\subsection{BoW}

The BoW model consisted of GloVe (300 dimension embeddings trained on 840B CommonCrawl tokens) \cite{pennington2014glove} vectors as the embedding layer.
The average of all word vectors for the input sequence is treated as its final representation. 
The representations for the hypothesis and premises were concatenated and passed through three fully connected layers with ReLU activation units after each layer. 
We concatenate the premise, hypothesis, their absolute difference and their product and pass it into the first layer of the classifier.
This input to the first layer is of 4 * embedding dimension and the output is of embedding dimension. 
Each subsequent hidden layer's input and output dimensions are embedding dimension * embedding dimension.

The model was trained with a vocabulary size of 50,000, a learning rate of 0.005, the maximum sequence length was 50 and a batch size of 32. 
We force all sentences to be of maximum sequence length using truncation or padding where applicable.
We train the model for 15 epochs and select the one that achieves highest validation accuracy for our experiments.

\subsection{BiLSTM}

The BiLSTM model consisted of GloVe (300 dimension embeddings trained on 840B CommonCrawl tokens) \cite{pennington-etal-2014-glove} vectors as the embedding layer.
The average of all word vectors for the input sequence is treated as its final representation.
The word vectors were passed through an LSTM unit. 
This unit was bidirectional, with 64 hidden units and 2 stacked LSTM layers.
The representations for the hypothesis and premises were concatenated and passed through three fully connected layers with ReLU activation units after each layer. 
We concatenate the premise, hypothesis, their absolute difference and their product and pass it into the first layer of the classifier. 
This input to the first layer is of hidden units * embedding dimension and the output is of embedding dimension. 
Each subsequent hidden layer's input and output dimensions are embedding dimension * embedding dimension.

The model was trained with a vocabulary size of 50,000, a learning rate of 0.005, the maximum sequence length was 50 and a batch size of 32. 
We force all sentences to be of maximum sequence length using truncation or padding where applicable.
We train the model for 15 epochs and select the one that achieves highest validation accuracy for our experiments.

\subsection{BERT}

We fine tuned the BERT-large model available on HuggingFace \footnote{\url{https://huggingface.co/bert-large-uncased}}. 
We added a classification head on top of the model using the AutoModel API on HuggingFace. 
The model was trained for 5 epochs with a learning rate of 3e-6 using the Adam optimizer. 
In order to simulate larger batch sizes on smaller GPUs, we used gradient accumulation as well. 
We simulated a batch-size of 32 by accumulating gradients over two batches of size 16. 
The model which achieved the highest validation accuracy was used for our experiments.

\subsection{RoBERTa}

We fine tuned the RoBERTa-large model available on HuggingFace \footnote{\url{https://huggingface.co/roberta-large}}. 
We added a classification head on top of the model using the AutoModel API on HuggingFace. 
The model was trained for 5 epochs with a learning rate of 3e-6 using the Adam optimizer. 
In order to simulate larger batch sizes on smaller GPUs, we used gradient accumulation as well. 
We simulated a batch-size of 32 by accumulating gradients over 8 batches of size 4. 
The model which achieved the highest validation accuracy was used for our experiments.

\subsection{GPT-3}

We used a temperature of 0.0 for all the experiments to select the most likely token at each step,
as this setting allow for reproducibility.

{\small
\begin{verbatim}
response = openai.Completion.create(
            model="text-davinci-003",
            prompt=prompt,
            temperature=0,
            max_tokens=1,
            top_p=1.0,
            frequency_penalty=0.1,
            presence_penalty=0.0
)
\end{verbatim}}

We restricted the model outputs to just one token.
Only ``yes" or ``no" are considered valid answers.
The model did not generate any output apart from these in all our experiments.
We used the following prompt template:

{\small
\begin{verbatim}
Premise: {sentence1}
Hypothesis: {sentence2}

Does the premise entail the hypothesis?
Answer:
\end{verbatim}}

\section{Dataset Creation}
\label{app:dataset-crowd}

The following process describes how we create a vetted, paraphrased version of the RTE dataset that tests whether models' are robust to paraphrased input.
First, we use a strong T5-based paraphraser to create three re-written sentences for each premise and hypothesis in the 2,400 pairs in the RTE1-3 test sets, resulting in 14,400 new sentences.
To generate these paraphrases, we use top-k sampling during decoding.\footnote{k=120; top-p=0.95} 
The re-writer model was fine-tuned on the Google PAWS dataset and can be found on Huggingface \footnote{\url{https://huggingface.co/Vamsi/T5_Paraphrase_Paws}}.
To evaluate its ability to generate gramatically correct paraphrases, we sampled 100 sentence pairs with at least one valid paraphrase and manually went through them.
Upon checking for grammaticality, we found a grammatical error in <8\% of the sentences.

Since we want to test paraphrastic understanding beyond simple lexical replacement, we discarded the re-written sentences that had at most a 25\% lexical overlap with the corresponding original sentence.
We use Jaccard index as a measure of lexical similarity \eqref{eqn:jaccard} where $\tau_s$ are the tokens in the original sentence and $\tau_p$ are the the tokens in the paraphrase.
\begin{equation}
\label{eqn:jaccard}
    Score = \frac{\tau_s \cap \tau_p}{\tau_s \cup \tau_p}
\end{equation}
To ensure that the re-written sentences are indeed sentence-level paraphrases for the original sentences, we relied on crowdsource workers to remove low quality paraphrases.
The Amazon Mechanical Turk HIT is described in detail in \autoref{appsubsec:mturk}.
We retain any paraphrases that get a similarity score above 75 out of 100.

\subsection{Manual Verification}
\label{appsubsec:pp_ver}

Before crowd sourcing to get the best paraphrase generated for a given sentence, we conducted manual evaluation to understand the average error rate of the paraphraser model used. As mentioned above, we sampled 100 sentence pairs with each pair having atleast one valid paraphrase. The paraphrases for these sentences were evaluated for grammatical errors. Any semantic errors are handled during crowd-sourcing. 

\begin{figure*}[t!]
    \centering
    \includegraphics[scale=0.5]
    {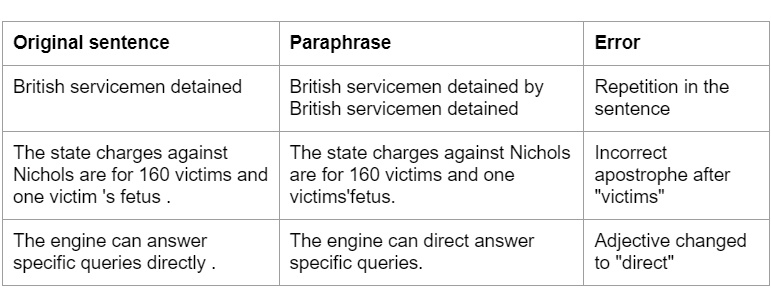}
    \caption{Types of errors made by the paraphraser model}
    \label{fig:paraphrase_errors}
\end{figure*}

The errors can roughly be classified into roughly three categories - repetition errors, tense errors and incorrect punctuation.
Examples of each type can be found in ~\autoref{fig:paraphrase_errors}.
Overall, we found the error rate to be small enough to continue using the paraphraser. We also asked MTurk workers to mark paraphrases as grammatically incorrect to ensure that the final dataset does not have any grammatically incorrect sentences.

\subsection{MTurk HIT}
\label{appsubsec:mturk}

We used Amazon Mechanical Turk to identify ungrammatical paraphrases rate how well a generated paraphrase preserved the meaning of the original sentence.
No filtering criteria was applied to crowdsource workers and were paid roughly \$14.20 an hour. 

\begin{figure*}[t!]
    \centering
    \includegraphics[scale=0.55]{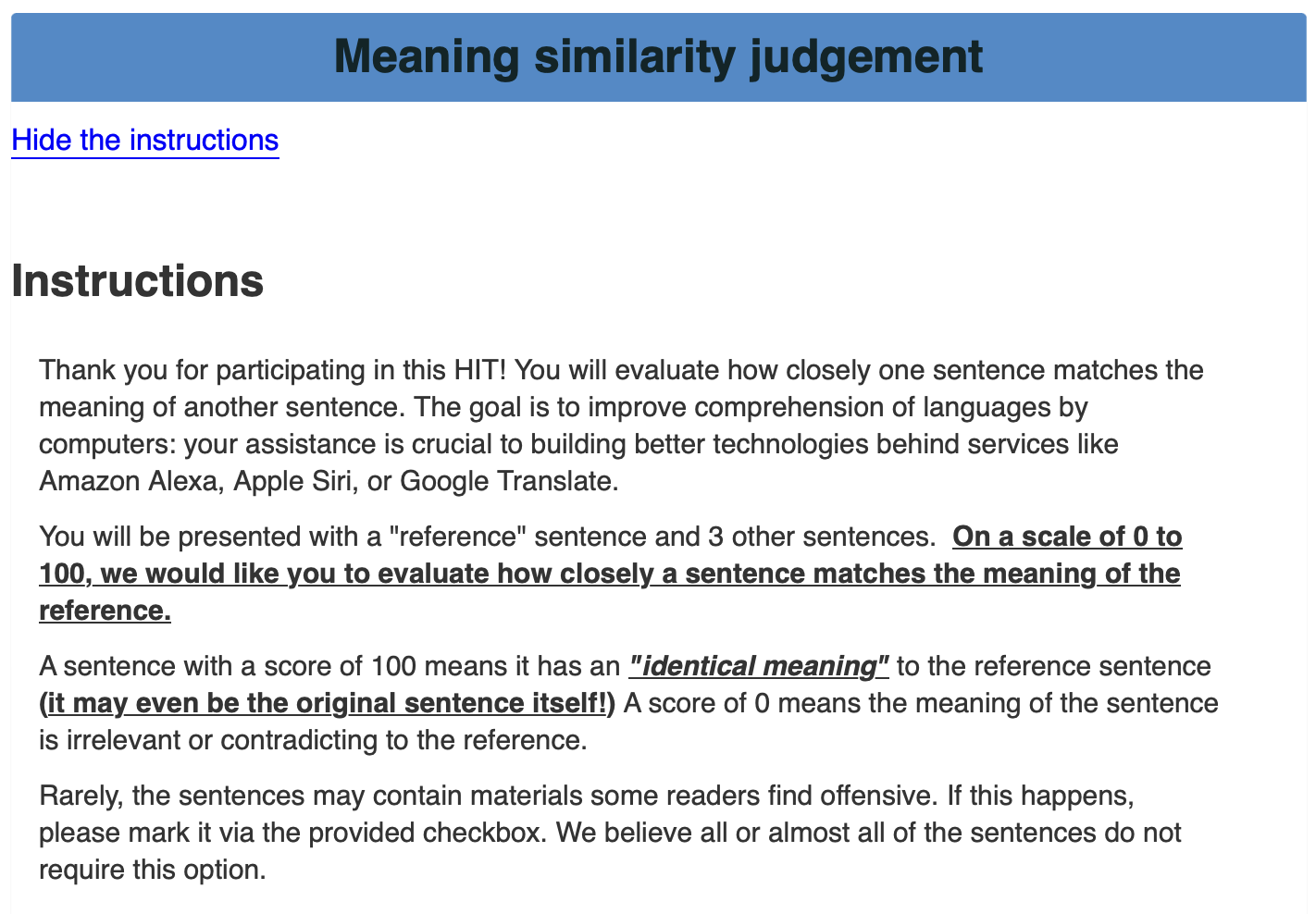}
    \caption{Instructions for semantic similarity and grammaticality check.}
    \label{fig:sim_annotate}
\end{figure*}

Each annotator was presented with a reference sentence,  a corresponding paraphrased sentences, and tasked to  
 judge on a scale of 0 to 100 how closely a paraphrased sentence retains the meaning of the reference sentence. 
A similarity score of 100 means that the paraphrase is the exactly the same in meaning as the reference, while a similarity score of 0 means that the meaning of the paraphrase is irrelevant or contradicts the reference sentence. 
Additionally, the MTurk workers were asked to judge the grammaticality of the paraphrase by selecting whether the paraphrase was grammatically correct or now. 
\autoref{fig:sim_annotate} includes the instructions we showed crowdsource workers for judging similarity between sentences.

\end{document}